 \documentclass[conference]{IEEEtran}

\usepackage{graphicx}
\usepackage{amsmath}
\usepackage{amssymb}
\usepackage[table]{xcolor}
 \usepackage{cite}
 
\usepackage{helvet}  
\usepackage{courier}  
\usepackage[hyphens]{url}  
\usepackage{graphicx} 
\usepackage{caption} 
\usepackage{amsmath}    
\usepackage{amsfonts}   
\usepackage{amssymb}    
\usepackage{mathtools}
\usepackage{colortbl}
\usepackage{multirow}
\UseRawInputEncoding
\usepackage{siunitx}
\usepackage{multirow}
\usepackage{xcolor,colortbl}
\usepackage{algorithm}
\usepackage{algorithmic}
\usepackage{graphicx}  
\usepackage{caption}   
\usepackage{newfloat}
\usepackage{listings}

\usepackage[colorlinks=true,
            linkcolor=blue,
            citecolor=blue,
            urlcolor=blue]{hyperref}
            
\IEEEoverridecommandlockouts                              

\title{\LARGE \bf
DETACH: Cross-domain Learning for Long-Horizon Tasks via \\ Mixture of Disentangled Experts
}

\author{
    Yutong Shen\textsuperscript{1}, 
    Hangxu Liu\textsuperscript{2}, 
    Lei Zhang\textsuperscript{3\dag}, 
    Penghui Liu\textsuperscript{1}, 
    Ruizhe Xia\textsuperscript{1}, 
    Tianyi Yao\textsuperscript{1}, 
    Tongtong Feng\textsuperscript{4\dag}
    \\
    {\small \textsuperscript{1}
    Beijing University of Technology, China}
    {\small \textsuperscript{2}
    Fudan University, China}
    {\small\textsuperscript{3}University of Hamburg, Germany}
    {\small\textsuperscript{4}Tsinghua University, China}\\
    {\small Corresponding authors:  fengtongtong@tsinghua.edu.cn, lei.zhang-1@studium.uni-hamburg.de}
}

\begin{document}

\maketitle
\thispagestyle{empty}
\pagestyle{empty}

\begin{abstract}

Long-Horizon (LH) tasks in Human-Scene Interaction (HSI) are complex multi-step tasks that require continuous planning, sequential decision-making, and extended execution across domains to achieve the final goal. However, existing methods heavily rely on skill chaining by concatenating pre-trained subtasks, with environment observations and self-state tightly coupled, lacking the ability to generalize to new combinations of environments and skills, failing to complete various LH tasks across domains. To solve this problem, this paper presents DETACH, a cross-domain learning framework for LH tasks via biologically inspired dual-stream disentanglement. Inspired by the brain's "where-what" dual pathway mechanism, DETACH comprises two core modules: i) an environment learning module for spatial understanding, which captures object functions, spatial relationships, and scene semantics, achieving cross-domain transfer through complete environment-self disentanglement; ii) a skill learning module for task execution, which processes self-state information including joint degrees of freedom and motor patterns, enabling cross-skill transfer through independent motor pattern encoding. We conducted extensive experiments on various LH tasks in HSI scenes. Compared with existing methods, DETACH can achieve an average subtasks success rate improvement of 23\% and average execution efficiency improvement of 29\%. More details can be found at: 
~\url{https://sites.google.com/view/detach-learning}.

\end{abstract}

\section{INTRODUCTION}


Long-Horizon (LH) tasks in Human-Scene Interaction (HSI) require continuous planning and cross-domain execution, posing challenges due to their complexity and need for environmental adaptation. These tasks have broad applications in robotics~\cite{qiu2024learning}, medical intervention~\cite{kim2024openvla}, and smart homes~\cite{kim2024openvla}, with canonical examples including dexterous hand manipulation~\cite{zhang2024contactdexnet} and humanoid whole-body control~\cite{sferrazza2024humanoidbench}. However, recent benchmarks show that HSI methods achieve low success rates on cross-domain tasks and demand extensive retraining~\cite{zhang2025interactanything,wang2024embodiedscan,xu2024interdreamer}, severely limiting real-world deployment.

\begin{figure}[!t]
    \centering
    \includegraphics[width=0.8\linewidth]{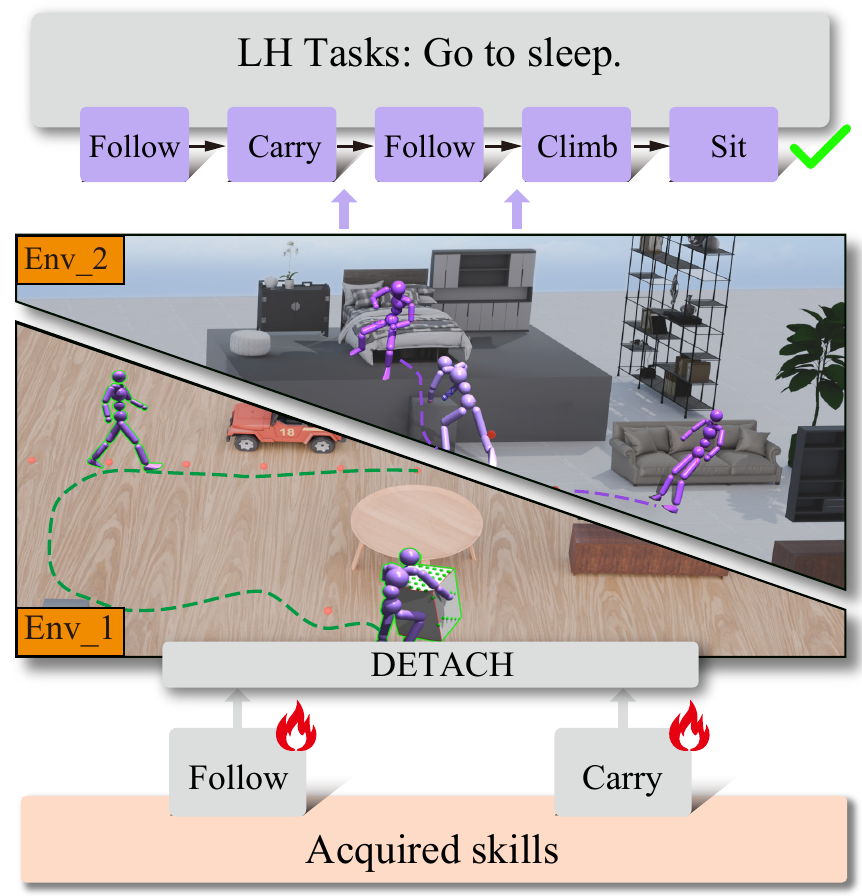}
    \caption{\small DETACH achieves generative generalization by learning fundamental subtasks in single environment (Env\_1), enabling it to generalize to novel environments and accomplish Long-Horizon tasks that involve previously unseen subtasks.}
    \label{fig:your-label}
\end{figure}

Recent large-scale vision-language-action (VLA) models~\cite{black2024pi0visionlanguageactionflowmodel,team2025gemini} and agent-based manipulation~\cite{ni2024don} 
achieve strong results on long-horizon embodied tasks. 
However, both paradigms typically adopt monolithic or tightly coupled end-to-end designs, where perception and control remain entangled, thereby limiting cross-domain generalization and modular skill reuse.

To bridge these gaps, current approaches \cite{pan2025tokenhsi,li2024optimus,park2024iclr} focus on processing self-state information in unified representation spaces, while other solutions  \cite{zhang2025interactanything,
xiao2023unified,
xu2024interdreamer
} 
further encode self-state information mixed with environmental information. 
The efficacy of the \textit{decompose-reuse-compose} paradigm has been confirmed by various studies \cite{huang2020one,
lan2023contrastive, xu2023composite,
hu2024disentangled
}, which also introduced a new modular learning paradigm for rapid adaptation to new skills by utilizing skill modules that have already been learned. 
In particular, CML~\cite{lan2023contrastive} and TokenHSI~\cite{pan2025tokenhsi} have explicitly demonstrated that such modular 
decomposition significantly outperforms standard end-to-end 
approaches in multi-task reinforcement learning (RL) and long-horizon 
HSI, respectively. 

Despite their promising performance, these methods suffer from the same architectural flaw: they adopt unified feature representation spaces that tightly couple environmental understanding with self-states. This flaw poses significant challenges in two main aspects: (1) Limited environmental transfer capability: When environmental changes occur (such as shifts from bright laboratory to dim factory settings), these systems cannot effectively separate the effects of environmental changes from self-state changes. This limitation necessitates relearning the entire perception-action mapping \cite{li2025controlling}, significantly constraining their cross-domain generalization capability. (2) Inefficient skill transfer capability: Current methods fail to achieve functional separation between perception and motor control. When encountering novel skills, even those involving similar motor patterns (such as grasping different objects), the system must retrain the entire perception-action network. This limitation makes it difficult to reuse, prevents effective reuse of learned motor skills, resulting in extremely low knowledge transfer efficiency due to a high risk of skill forgetting \cite{van2024continual}. Even
advanced modular approaches such as CML~\cite{lan2023contrastive} 
and TokenHSI~\cite{pan2025tokenhsi} still rely partly on unified feature spaces, thereby inheriting some of these limitations.

To address these challenges, this paper introduces \textbf{DETACH}: a biologically inspired functional disentanglement architecture that draws from the dorsal-ventral stream hypothesis in neuroscience \cite{ungerleider1982two}. According to this hypothesis, the brain's ventral \textit{what} pathway specializes in object recognition, while the dorsal \textit{where-how} pathway handles spatial processing and motor control. Unlike existing dual-stream approaches \cite{ibrayev2024toward} that separate visual modalities, DETACH introduces a functional disentanglement: the \textbf{Environmental Encoder} learns scene-invariant spatial relationships \cite{arkhangelsky2024causal} while the \textbf{Self-Encoder} captures body-schema-specific motor primitives.

Proposed method is extensively evaluated on various self-designed LH-embodied AI tasks, including cross-scene adaptation, novel skill adaptation, and particularly LH control tasks in complex environments. The contributions of this paper can be summarized as follows.

\begin{itemize}
\item Proposing the \textbf{DETACH disentangled architecture}, the first Embodied AI control framework in HSI based on biologically inspired cognitive principles. This architecture separates traditional unified encoding into specialized parallel processing of environmental perception streams and self-state perception streams.

\item Designing \textbf{specialized dual-stream encoders}, where the environmental encoder enhances \textbf{cross-domain transfer capability}, and the self-encoder achieves \textbf{cross-task skill reuse}. Both encoders are independently optimized and flexibly combined.

\item Establishing comprehensive benchmark scenarios for LH tasks through designed progressive LH task benchmarks, and validating the effectiveness of DETACH on these benchmarks. Compared to existing methods, DETACH achieves a \textbf{$2\times$ improvement in cross-domain adaptation capability} and a \textbf{$1.5\times$ improvement in skill reuse efficiency}.
\end{itemize}

\section{Related Works}
\subsection{Human-Scene Interaction}

HSI focuses on enabling embodied agents to interact naturally and effectively with complex 3D environments. Existing approaches include unified representation learning (e.g., Chain of Contacts \cite{xiao2023unified}), which integrates contact and object encoding with LLM-based planning but exhibits limited generalizability due to tight coupling between perception and action components; staged processing (e.g., Dynamic HSI~\cite{jiang2024scaling}), which uses autoregressive diffusion for disentangled scene understanding and action generation, ensuring temporal coherence but at high computational cost; and end-to-end methods (e.g., TokenHSI~\cite{pan2025tokenhsi}, and \cite{zhang2025interactanything,xu2024interdreamer}), which synthesize motion from text using pre-trained models and object sensors but are limited to simple skill composition scenarios. A key limitation shared by these approaches is their tight coupling between perception and control modules, which hinders cross-domain transfer and skill reusability.

\begin{figure*}[h]  
  \centering
  \includegraphics[width=0.9\textwidth]{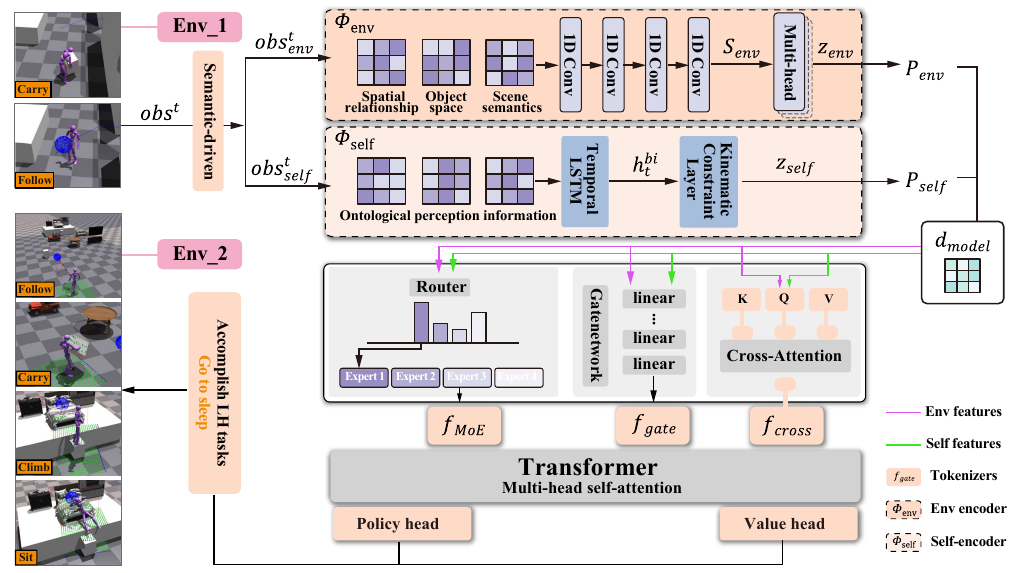}  
  \caption{Illustrating the operational workflow of the DETACH, Raw observation $\text{obs}^t$ is semantically disentangled into environmental $\text{obs}^t_{\text{env}}$ and self-state $\text{obs}^t_{\text{self}}$ components. Environmental encoder $\Phi_{\text{env}}$ and self-encoder $\Phi_{\text{self}}$ process respective inputs, with projection layers $P_{\text{env}}$ and $P_{\text{self}}$ mapping outputs to unified $d_{\text{model}}$ space. Multi-strategy adaptive fusion integrates features via three components: MoE fusion, gated fusion network, and cross-attention fusion module, producing outputs $(f_{\text{MoE}}, f_{\text{gate}}, f_{\text{cross}})$. These fused representations undergo Transformer multi-head self-attention before feeding into policy and value heads. }  
  \label{2}  
\end{figure*}
\subsection{Long-Horizon Task}
LH tasks in HSI require agents to perform multi-step reasoning and manage long-term dependencies \cite{li2024optimus}. Current approaches include hierarchical planning (e.g., MLLM-based instruction parsing with visual encoders \cite{li2025optimus2,zheng2023steve}), which decomposes tasks into subgoals but suffers from low skill prediction accuracy; memory augmentation (e.g., hierarchical memory and knowledge graphs \cite{li2024optimus}), which models long-term dependencies yet lack dynamic adaptation; and causal modeling \cite{li2025optimus3}, which enhances policy learning through observation-action causality but requires high computational resources and relies on limited training data. These methods are limited by their reliance on static representations, which constrains cross-domain transfer, policy reuse, and adaptation to dynamic interaction scenarios.

\subsection{Disentangled Learning}

Disentangled representation learning addresses these limitations by decomposing complex systems into independent, interpretable modules, improving generalization and controllability \cite{ada2024diffusion}. Key approaches include mutual information-based disentanglement (e.g., \cite{hu2024disentangled}), which minimizes mutual information between skill components but requires domain-specific prior knowledge; factorized representation learning (e.g., $\beta$-VAE framework \cite{uppal2025denoising}), which uses disentanglement regularization; and variational disentanglement (e.g., \cite{bhowal2024variational}), which optimizes a variational lower bound. An alternative method \cite{yang2025task} employs Wasserstein distance for stable disentanglement, though it remains theoretical, while \cite{yang2025task} also identifies valuable factors at high computational cost. However, these methods focus on static factor separation, which are ill-suited for the dynamic, continuous interactions and generative adaptation required in LH embodied tasks.

\section{Methods}

DETACH employs a dual-encoder design with environmental encoder $\Phi_{\text{env}}$ and self-encoder $\Phi_{\text{self}}$ to disentangle environmental perception from self-state representation. Their outputs are fused via a multi-strategy adaptive mechanism and processed by the shared transformer encoder $\phi$ to enhance perception-control collaboration.

\subsection{ Observation Space Reconstruction Model}

DETACH employs observation disentanglement, modeling unified observation space as a Dual-Stream Separation Process (DSP). The disentanglement objective minimizes mutual information between environmental and self-state representations, quantified as $D = \sum_{t=0}^{T} \gamma^t I(obs^t_{env}, obs^t_{self})$, implemented using correlation-based mutual information estimators.

\subsection{ Disentangled Dual-Encoder}
DETACH is a biologically inspired, disentangled dual-encoder architecture that separates the traditional unified encoding pathway into two specialized processing streams for environmental perception and self-state representation. Figure 2 illustrates the proposed disentanglement module, which consists of four key components:

\textit{Environmental encoder $\Phi_{env}$.} The environmental encoder processes spatial information such as object positions and scene semantics. Given environmental observations $obs^t_{env} \in \mathbb{R}^{T \times d_{env}}$, we adopt parallel convolutional layers for feature extraction. Feature extraction is performed as:
\begin{equation}
S_{env} = \text{Concat}[\text{Conv1D}_k(obs^t_{env})]
\end{equation}

Features are aggregated through multi-head self-attention for spatial feature aggregation:
\begin{equation}
\begin{split}
z_{env} &= \text{LayerNorm}(\text{MultiHeadAttn}(S_{env}, S_{env}, S_{env}) \\ &\quad + S_{env})
\end{split}
\end{equation}

The environmental encoder is paired with decoder $\text{Decoder}_{env}$ for reconstruction-based pre-training.

\textit{Self-encoder $\Phi_{self}$.} The self-encoder processes self-state information $obs^t_{self} \in \mathbb{R}^{T \times d_{self}}$ including joint angles and velocities. Since accurate self-state understanding requires bidirectional temporal context for accurate motion understanding, the self-encoder employs a recurrent neural architecture with bidirectional processing capabilities. The model is defined as:
\begin{equation}
h^{bi}_t = [h^f_t; h^b_t]
\end{equation}
where $h^f_t$ and $h^b_t$ represent forward and backward temporal representations, respectively.

The kinematic constraint layer ensures outputs remain within physically feasible ranges through a element-wise soft gating mechanism:
\begin{equation}
z_{self} = h^{bi}_t \odot \sigma(W_k h^{bi}_t + b_k)
\end{equation}
where $\sigma$ is the sigmoid function, and $W_k$ and $b_k$ are learnable parameters. The kinematic constraint layer ensures the physical feasibility of generated actions.

The self-encoder is paired with a temporal prediction network $f_{pred}$ for sequence prediction-based pre-training, which learns to predict future self-state representations from current ones.
\textit{Feature projection layers $P_{env}$ and $P_{self}$.} Two independent linear layers map the encoder outputs to a unified $d_{model}$ dimensional space:
\begin{equation}
f_{env} = P_{env}(z_{env}), \quad f_{self} = P_{self}(z_{self})
\end{equation}
where $f_{env}, f_{self} \in \mathbb{R}^{d_{model}}$ are the projected features used for fusion.

\subsection{Multi-Strategy Adaptive Fusion Mechanism}
To effectively integrate heterogeneous features from the environmental encoder and self-encoder, \textbf{DETACH} incorporates a multi-strategy adaptive fusion mechanism that combines three complementary fusion strategies. According to Figure 2, the adopted fusion mechanism comprises three core components:

\textit{Cross-attention fusion module.} This module \cite{vaswani2017attention} is selected for its capability to enable internal states to actively query key information from the environment, thereby achieving state-driven dynamic feature alignment. It uses self-state features $f_{self} \in \mathbb{R}^{d_{model}}$ as Query, and environment features $f_{env} \in \mathbb{R}^{d_{model}}$ as Key and Value, achieving dynamic weight allocation through a multi-head attention mechanism:

\begin{equation}
\begin{split}
f_{cross} &= \text{MultiHead}(f_{self}, f_{env}, f_{env}) \\
&= \text{Concat}(\text{head}_1, \ldots, \text{head}_h)W^O
\end{split}
\end{equation}
where each attention head:
\begin{equation}
\text{head}_i = \text{Attention}(f_{self} W^Q_i, f_{env} W^K_i, f_{env} W^V_i)
\end{equation}


\textit{Gated Fusion Network.} This module dynamically modulates contribution weights between environmental perception and self-state features to prevent imbalance, using learnable gating units. It is implemented via a multi-layer MLP \cite{tolstikhin2021mlp} with decreasing hidden units, matching the fused feature dimension. The gated fusion strategy is defined as:
\begin{multline}
f_{gate} = \sigma(W_g [f_{env}; f_{self}] + b_g) \odot f_{env} \\
+ (1 - \sigma(W_g [f_{env}; f_{self}] + b_g)) \odot f_{self}
\end{multline}

\textit{Mixture of Experts (MoE) fusion module .} This module is adopted for its capacity to dynamically select optimal fusion experts based on task characteristics and environmental complexity, enabling adaptive feature integration. It designs multiple specialized fusion experts \cite{zadouri2023pushing}, each modeled by a multi-layer MLP network with a hierarchical structure, dynamically selecting the most suitable expert for feature fusion through a routing network. The mixture of experts' fusion is represented as:
\begin{equation}
f_{moe} = \sum_{i=1}^{4} w_i \cdot E_i(f_{env}, f_{self})
\end{equation}
where the routing weights are
\begin{equation}
w_i = \text{Softmax}(W_r [f_{env}; f_{self}] + b_r)_i
\end{equation}

The three fusion strategies are combined through a learnable weighted combination:
\begin{equation}
f_{fused} = \alpha \cdot f_{cross} + \beta \cdot f_{gate} + \gamma \cdot f_{moe}
\end{equation}
where $\alpha, \beta, \gamma$ are learnable parameters that balance the contributions of different fusion strategies.

\textit{Shared Transformer Encoder.} The fused features are processed by a shared transformer encoder $\phi$ to enhance perception-control collaboration:
\begin{equation}
h_{transformer} = \phi(f_{fused})
\end{equation}
where $\phi$ consists of multiple transformer layers with self-attention mechanisms to capture long-range dependencies and temporal relationships.

\textit{Policy and Value Heads.} The transformer output is fed into separate policy and value heads for action prediction and value estimation:
\begin{equation}
\begin{split}
\pi(a|s) &= \text{PolicyHead}(h_{transformer}) \\
V(s) &= \text{ValueHead}(h_{transformer})
\end{split}
\end{equation}

where $\pi(a|s)$ represents the action probability distribution and $V(s)$ represents the state value function.

\subsection{ Progressive Training Protocol}
To fully leverage the advantages of disentangled architecture and ensure the specialized characteristics of each module, DETACH designed a comprehensive progressive training protocol and specialized regularization mechanisms. The adopted training protocol involves three progressive stages, each with clear training objectives and parameter update strategies:

\textit{Independent Pre-training Stage.} In this stage, the environmental encoder $\Phi_{env}$ and self-encoder $\Phi_{self}$ are trained independently to establish their respective feature representation capabilities. The environmental encoder is pre-trained through the scene reconstruction loss:
\begin{equation}
\mathcal{L}_{env} = \|\text{Decoder}_{env}(\Phi_{env}(obs^t_{env})) - obs^t_{env}\|_2^2
\end{equation}

The self-encoder is pre-trained through action sequence prediction tasks:
\begin{equation}
\mathcal{L}_{self} = \sum_{t=1}^{T-1} \|\Phi_{self}(obs^{t+1}_{self}) - f_{pred}(\Phi_{self}(obs^t_{self}))\|_2^2
\end{equation}
where $f_{pred}$ is the temporal prediction network.This stage establishes domain-specific representation foundations.

\textit{Fusion Layer Optimization Stage.} In this stage, the pre-trained encoder parameters $\theta_{env}, \theta_{self}$ are frozen to preserve learned representations, focusing on training the feature fusion layer and Transformer encoder $\phi$:
\begin{equation}
\mathcal{L}_{fusion} = \mathcal{L}_{task} + \lambda_{quality} \mathcal{L}_{fusion\_quality}
\end{equation}
where $\mathcal{L}_{task}$ represents the standard reinforcement learning objective (e.g., policy gradient loss for PPO), which guides the agent to maximize expected cumulative rewards.

where the fusion quality loss is defined as:
\begin{equation}
\begin{split}
\mathcal{L}_{fusion\_quality} &= \|f_{cross} - (f_{env} + f_{self})\|_2^2 \\
&\quad + \lambda_{disentangle} \cdot I(z_{env}, z_{self})
\end{split}
\end{equation}
where $I(z_{env}, z_{self})$ represents the mutual information between environmental and self-state features. The first term ensures fusion consistency, while the second term maintains disentanglement by minimizing mutual information between representations.

\textit{End-to-End Joint Optimization Stage.} In the end-to-end joint optimization stage, all network parameters are unfrozen for end-to-end joint optimization, while introducing specialized preservation regularization:
\begin{equation}
\begin{split}
\mathcal{L}_{total} &= \mathcal{L}_{task} + \lambda_{disentangle} \cdot I(z_{env}, z_{self}) \\
&\quad + \sum_{i} \lambda_i \mathcal{R}_i
\end{split}
\end{equation}
where the regularization terms include:
\begin{align}
\mathcal{R}_1 &= \|\theta_{env} - \theta_{env}^*\|_2^2 \quad \text{(encoder preservation)} \\
\mathcal{R}_2 &= \|\theta_{self} - \theta_{self}^*\|_2^2 \quad \text{(encoder preservation)} \\
\mathcal{R}_3 &= \sum_{i \neq j} \|f_i - f_j\|_2^2 \quad \text{(fusion diversity)}
\end{align}
where $f_i, f_j \in \{f_{cross}, f_{gate}, f_{moe}\}$ and $\mathcal{R}_3$ encourages different fusion strategies to learn complementary representations. The regularization weights $\lambda_i$ (where $i \in \{1,2,3\}$) are hyperparameters that balance the contributions of different regularization terms. $\theta_{env}^*$ and $\theta_{self}^*$ are the pre-trained encoder parameters, and the disentanglement regularization term $I(z_{env}, z_{self})$ ensures continued minimization of mutual information between environmental and self-state representations throughout the training process.

\begin{figure}[t]
    \centering
    \includegraphics[width=0.45\textwidth]{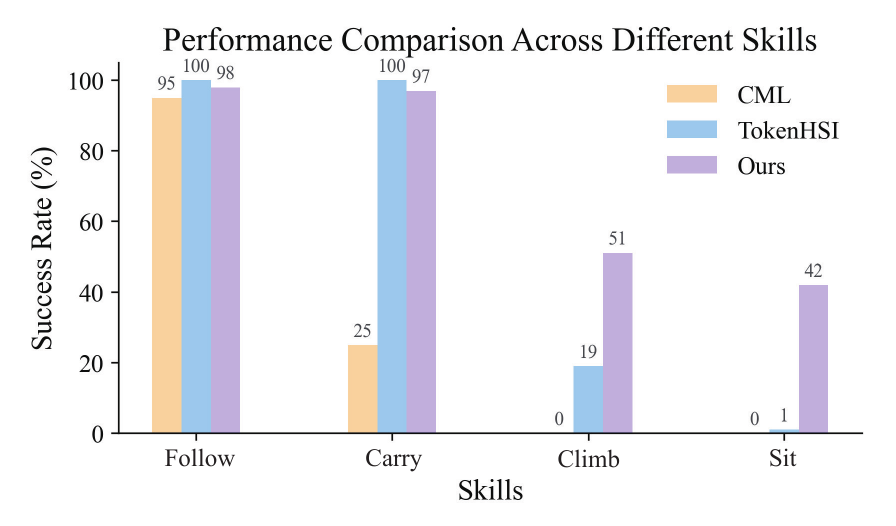}
    \caption{Success rate comparison across different skills among CML, TokenHSI, and our DETACH framework.}
    \label{fig:skill_comparison}
\end{figure}
\begin{table}[t]
\centering
\renewcommand{\arraystretch}{0.65} 
\begin{tabular}{cccccc}
\hline
\rule{0pt}{3ex}\textbf{Method} & \textbf{Follow} & \textbf{Carry} & \textbf{Climb} & \textbf{Sit} & \textbf{LH1} \\[1ex]
\hline
\rule{0pt}{2.5ex}CML~\cite{xu2023composite} & 0.95 & 0.25 & 0.00 & 0.00 & 0.30 \\[0.5ex]
TokenHSI~\cite{pan2025tokenhsi} & 1.00 & 1.00 & 0.19 & 0.01 & 0.55 \\[0.5ex]
Ours & 0.98 & 0.97 & \cellcolor{gray!20}\textbf{0.51} & \cellcolor{gray!20}\textbf{0.42} & \cellcolor{gray!20}\textbf{0.72} \\[0.5ex]
\hline
\end{tabular}
\caption{Success rates for foundational skills and composite task completion.}
\end{table}
\begin{table*}[htbp]
\centering
\renewcommand{\arraystretch}{1.3}  
\resizebox{1.0\textwidth}{!}{
\begin{tabular}{c|c S[table-format=1.2] S[table-format=1.2] S[table-format=1.2] S[table-format=1.2] S[table-format=1.2] S[table-format=3.2] S[table-format=1.2]|S[table-format=1.2]|S[table-format=1.2]}
\hline
\textbf{Experiment} & \textbf{Method} & \textbf{Follow} & \textbf{Carry} & \textbf{Follow} & \textbf{Climb} & \textbf{Sit} & \textbf{Time(s)} & \textbf{LH.} & \textbf{SGR.} & \textbf{EGR.} \\
\hline
\multirow{2}{*}{LH2} & TokenHSI & 1.00 & 0.56 & 0.13 & {-} & 0.01 & 99.00 & 0.42 & 0.01 & 0.76 \\
 & \cellcolor{gray!20} Ours & \cellcolor{gray!20} \textbf{1.00} & \cellcolor{gray!20} \textbf{0.96} & \cellcolor{gray!20} \textbf{0.67} & \cellcolor{gray!20} \textbf{-} & \cellcolor{gray!20} \textbf{0.16} & \cellcolor{gray!20} \textbf{85.00} & \cellcolor{gray!20} \textbf{0.70} & \cellcolor{gray!20} \textbf{0.08} & \cellcolor{gray!20} \textbf{0.97} \\
\hline
\multirow{2}{*}{LH3} & TokenHSI & 1.00 & 0.50 & 0.21 & 0.20 & 0.00 & 102.90 & 0.38 & 0.67 & 0.69 \\
 & \cellcolor{gray!20} Ours & \cellcolor{gray!20} \textbf{1.00} & \cellcolor{gray!20} \textbf{0.95} & \cellcolor{gray!20} \textbf{0.50} & \cellcolor{gray!20} \textbf{0.40} & \cellcolor{gray!20} \textbf{0.10} & \cellcolor{gray!20} \textbf{97.60} & \cellcolor{gray!20} \textbf{0.59} & \cellcolor{gray!20} \textbf{0.13} & \cellcolor{gray!20} \textbf{0.81} \\
\hline
\end{tabular}
}
\caption{Comparison of generalization performance between TokenHSI and DETACH on LH tasks.}
\label{tab:experiment_results}
\end{table*}

\section{Experiment}
We conduct comprehensive experiments to evaluate our method across foundational skill learning and Long-Horizon(LH) task execution. Section IV-A assesses the robustness of foundational skill acquisition and task completion. Section IV-B provides ablation studies on each component of the framework. Section IV-C illustrates the model's generalization to complex LH tasks with diverse skill and environment compositions.

Our experiments are conducted entirely on three LH tasks that we designed:

\textbf{LH1: ``Sit on Chair!''} This LH task comprises a sequence of four fundamental skills: \textit{Follow, Carry, Climb,} and \textit{Sit}, where the target object for \textit{Sit} is a Chair.

\textbf{LH2: ``Sit on Sofa!''} This LH task similarly comprises a sequence of four fundamental skills: \textit{Follow, Carry, Follow,} and \textit{Sit}, where the target object for \textit{Sit} is a Sofa.

\textbf{LH3: ``Go to Bed!''} This LH task comprises a sequence of five fundamental skills: \textit{Follow, Carry, Follow, Climb,} and \textit{Sit}, where the target object for \textit{Sit} is a Bed.

Our object assets are sourced from the 3D-FRONT dataset \cite{fu20213d}, while the motion data is inherited from TokenHSI~\cite{pan2025tokenhsi}.
\begin{figure}[t]
    \centering
    \includegraphics[width=0.4\textwidth]{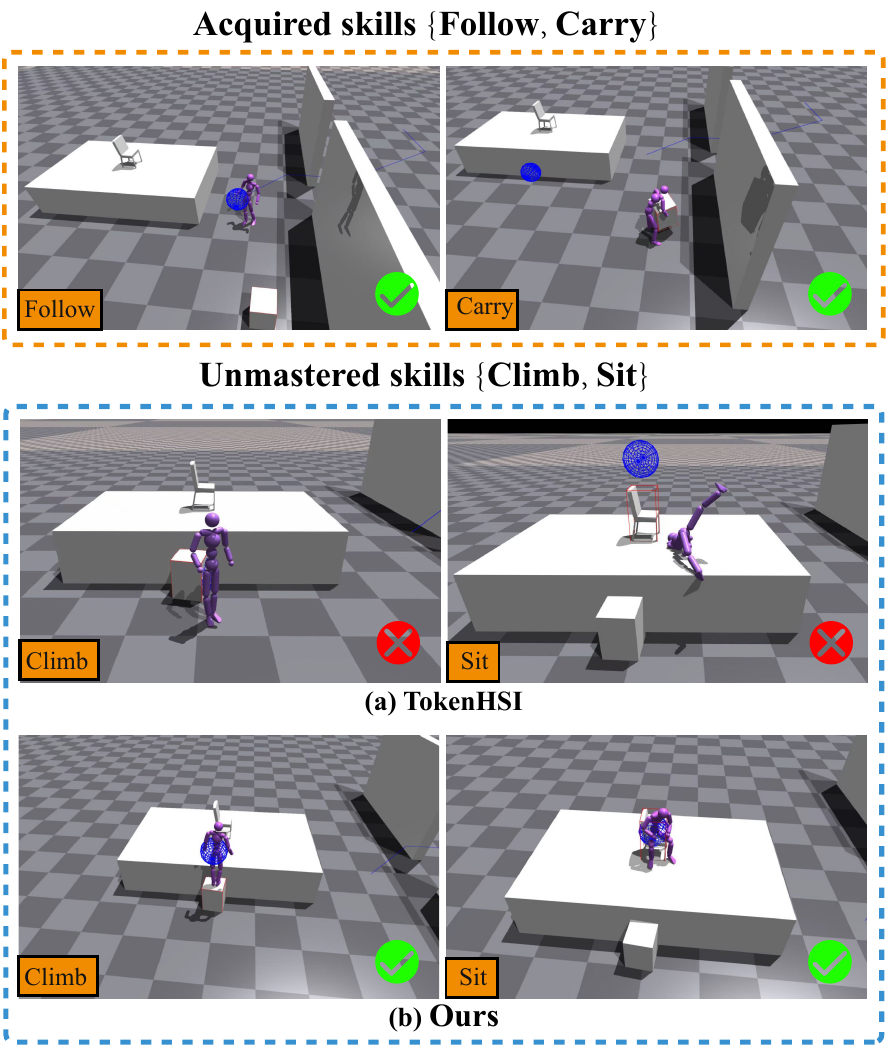}
    \caption{Skill acquisition performance comparison between Detach and TokenHSI. The orange box represents the skills learned in pre-training, and the blue boxes represent new generalized skills.}
    \label{fig:skill_acquisition}
\end{figure}

\subsection{ Evaluation on Foundational Skill Learning and Task Completion}
\textbf{Experimental Setup.}  
To evaluate the robustness and universality of our disentangled architecture, we employ a progressive learning protocol where foundational skills \textit{Follow} and \textit{Carry} are established through comprehensive training, while \textit{Climb} and \textit{Sit} skills are acquired through compositional learning. This approach enables systematic assessment of skill generalization capabilities and adaptation to diverse environments.
The training procedure uses large-scale parallelization in 4,096 environments, employing PPO \cite{schulman2017proximal} with 10k iterative updates. We conducted 100 independent experimental trials to ensure statistical reliability, quantifying robustness and universality through the success rate means of all skills and L1 tasks. This rigorous evaluation framework provides a comprehensive assessment of the architecture's performance through systematic skill composition and environmental adaptation.

\begin{figure*}[t]  
  \centering
  \includegraphics[width=0.90\textwidth]{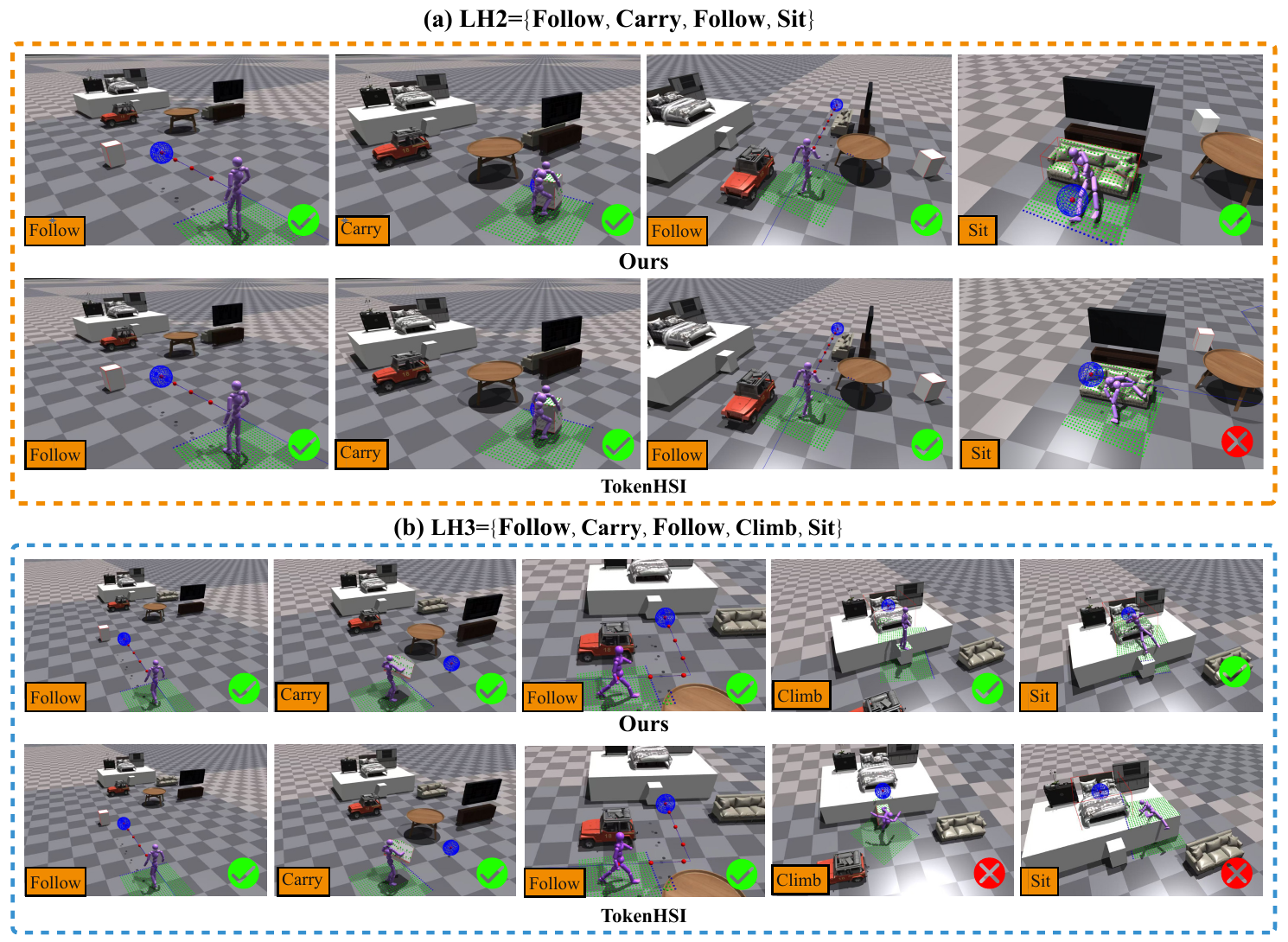}  
  \caption{Generalization comparison between DETACH and TokenHSI on LH tasks, where (a) and (b) represent tasks composed of sequences of four and five foundational skills, respectively. We only pre-trained the first two actions, \textit{Follow} and \textit{carry} on LH1 tasks, and tested skill generalization and environmental generalization in new scenarios.}  
  \label{2}  
\end{figure*}

\textbf{Baselines.} 
We train TokenHSI from scratch using our custom dataset. TokenHSI is a state-of-the-art full-body humanoid controller that learns a set of foundational skills comparable to ours. We also include CML~\cite{xu2023composite}, a composite motion learning baseline commonly used alongside TokenHSI, as an additional point of reference.

\textbf{Follow and Carry.} The success rate of \textit{Follow} is defined as maintaining the pelvis within a 30cm distance threshold from the target path in the XY plane. For the \textit{Carry} task, which can be decomposed into 'grasp' and 'transport' components, achieving only the grasp phase without successful transport to the designated target location is considered 0.5 task completion. \textit{Follow} task training utilized procedurally generated trajectories, while \textit{Carry} task training employed 9 boxes of varying dimensions. Subsequently, we trained on the compositional LH1 task combining these two primitives and evaluated performance on identical task compositions.

\textbf{Climb and Sit.} The success of \textit{Climb} is defined as reaching the target object with the pelvis positioned at or above the target elevation. Success of \textit{Sit} requires the pelvis to be positioned on the upper surface of the target object.

\textbf{LH1 task.} The Success rate for LH1 is the success rate of the sub-skill sequence. Due to the sequential nature of LH tasks, where skills must be executed in order, failure in a preceding task prevents the execution of subsequent tasks. Therefore, the skill sequence success rate serves as an excellent metric for evaluating the success rate of LH tasks.

\textbf{Results.}
Table I presents the quantitative analysis results, where we evaluated the effectiveness of three methods: CML, TokenHSI, and DETACH. While all methods demonstrate comparable performance on pre-trained skills such as \textit{Follow} and \textit{Carry}, Figure 3 and  Figure 4 reveal that, compared to methods with limited generalization capabilities like CML and TokenHSI, our DETACH method maintains high success rates for pre-trained skills while achieving success rates of 51\% and 42\% on two additional tasks, \textit{Climb} and \textit{Sit}, respectively, significantly outperforming the other two approaches. In contrast, TokenHSI and CML exhibit limited generalization on these tasks, resulting in success rates approaching zero. Furthermore, in terms of overall task success rate, DETACH achieves 72\%, surpassing CML and TokenHSI by 42\% and 17\%, respectively. These results highlight DETACH's stability in executing existing skills and its versatility in handling novel tasks, demonstrating its superior performance capabilities.

\subsection{Long-Horizon Task Completion}

This section evaluates the DETACH framework's performance on Long-Horizon (LH) tasks, designed to test generalization across skills and environments. We focus on \textbf{skill generalization} and \textbf{environment generalization}, using LH2 and LH3 for assessment, which target adaptation to novel environments and task compositions. Generalization is evaluated over 100 test runs per task, measuring subtask success rates in diverse, unseen scenes to assess robustness.

\textbf{Task Execution Times.} Task execution time refers to the duration from the start of the current LH task to the initiation of the next LH task. The criteria for determining the execution of the next task include the occurrence of errors (such as falling) or exceeding the threshold time for task execution.

\textbf{Experiment setup.}
As described in Section IV-A, to validate the environment generalization capability in this section, we employ the same progressive learning protocol on the LH1 task and directly evaluate generalization on the LH2 and LH3 tasks. This approach allows us to observe the environment generalization capability of our DETACH framework more intuitively. Since we similarly establish foundational skills \textit{Follow} and \textit{Carry} through comprehensive training, we can also assess skill generalization rates through the completion performance on \textit{Climb} and \textit{Sit}.

\textbf{Generalization Rate Definition.} Based on our experimental data, we formally define the Environment Generalization Rate (EGR) and Skill Generalization Rate (SGR) as follows:
\begin{equation}
EGR = \frac{S_{Li}}{S_{L1}}, i \in {2,3}
\label{eq:egr}
\end{equation}

\begin{equation}
SGR = \frac{(S_{climb} + S_{sit})/2}{(S_{follow} + S_{carry})/2}
\label{eq:sgr}
\end{equation}
where $S_{Li}, i \in \{1,2,3\}$ represents the success rate of LH tasks, and the testing on LH2 and LH3 involves direct transfer from the LH1 environment training, demonstrating its rationality. Similarly, $S_{climb}$ etc. represent the success rates of skills, where \textit{Climb} and \textit{Sit} are composed from foundational \textit{Follow} and \textit{Carry} skills; therefore, we define the skill generalization rate using this formula.

\textbf{Results.}
Figure 5 visually highlights DETACH's superior skill composition over TokenHSI. Table II compares subtask success rates, LH task success rates, execution times, and environment/skill generalization rates, showing DETACH's consistent outperformance. Specifically, DETACH reduces average execution times by 14s (LH2) and 5s (LH3) compared to TokenHSI, enhancing efficiency in human body control. Task success rates improve by 28\% (LH2) and 21\% (LH3), balancing efficiency with success. Notably, environment generalization rates reach 0.08 (LH2) and 0.13 (LH3), with skill generalization rates of 0.97 (LH2) and 0.81 (LH3), significantly exceeding TokenHSI. These results underscore DETACH's enhanced composition capabilities for long-horizon HSI tasks.

\begin{table}[t] 
\small 
\setlength{\tabcolsep}{4pt} 
\renewcommand{\arraystretch}{1.2}
\centering
\resizebox{0.47\textwidth}{!}{
\begin{tabular}{c|l|ccc}
\hline
\textbf{ID} & \textbf{Configuration} & \textbf{EGR.} & \textbf{SGR.} & \textbf{LH.} \\
\hline
\textbf{Full} & All modules enabled & \textbf{0.81} & \textbf{0.13} & \textbf{0.58} \\
A1 & w/o Env Encoder $\Phi_{\text{env}}$ & \cellcolor{gray!20}\textbf{0.64} & 0.12 & \cellcolor{gray!20}\textbf{0.41} \\
A2 & w/o Self Encoder $\Phi_{\text{self}}$ & 0.74 & \cellcolor{gray!20}\textbf{0.05} & \cellcolor{gray!20}\textbf{0.38} \\
\hline
\end{tabular}
}
\caption{Ablation study results. Each variant disables one key module. Bold indicates best performance.}
\label{tab:ablation}
\end{table}

\subsection{ Ablation Experiment}

To evaluate the individual contributions of key modules in our DETACH framework, we perform a comprehensive ablation study. Each variant is constructed by disabling or removing a specific component from the full model while keeping all other settings fixed. The experiments are conducted on  LH3 tasks composed of foundational skill primitives (e.g., \textit{follow}, \textit{carry}, \textit{climb}, \textit{sit}) in diverse environments.

\textbf{Experimental Setup.}We recorded three key metrics: environment generalization success rate, skill generalization success rate, and overall LH task success rate. Each model was trained under the same progressive learning protocol as described in Section IV-A and evaluated by executing LH3, with results shown in Table III below.

\textbf{Results.}
As shown in Table III, removing the environmental encoder (\textbf{A1}) causes the environment generalization rate to drop from 0.81 to 0.64, while removing the self encoder (\textbf{A2}) leads to a decrease in skill generalization rate from 0.127 to 0.045. Removing any component results in either substantial or moderate degradation across all metrics. This confirms that each encoder in our framework serves a distinct function and is indispensable to the overall architecture.

\section{Conclusion and Future Work}
In this work, we presented \textbf{DETACH}, a biologically inspired dual-stream disentanglement framework that explicitly separates environment understanding from self-state encoding. This design enables \textbf{cross-domain transfer, modular skill reuse, and efficient long-horizon task composition}.
Extensive experiments on diverse HSI scenarios demonstrate that DETACH achieves substantial improvements of 23\% in subtask success rate and 29\% in execution efficiency, along with stronger generalization over state-of-the-art modular baselines.

While our current implementation relies on a pre-defined skill set, future work will explore open-ended skill discovery from unlabeled data and real-world deployment under dynamic environments. We believe DETACH provides a promising step toward scalable, generalizable embodied intelligence in complex human-scene interactions.



\vspace{0.1in}

\bibliographystyle{IEEEtran}
\bibliography{IEEEabrv,root}

\newpage
\appendix
\subsection{Simulated Charactor}
Our humanoid dataset builds upon TokenHSI \cite{pan2025tokenhsi}.

TokenHSI develops a customized character model with 32 degrees of freedom based on the AMP 
system, comprising 15 rigid bodies, 12 controllable joints, and 32 degrees of freedom. To address the mismatch between the SMPL  parameters used in reference motion datasets and the kinematic structure of the AMP character model, TokenHSI implements three key improvements:
\begin{enumerate}
    \item Adjustment of the 3D positions of lower limb joints (hip, knee, and ankle joints) to align with the SMPL human body model;
    \item Adoption the SimPoE method to transform foot collision shapes from rectangular boxes to realistic foot meshes;
    \item Upgrading knee joints from 1-DOF rotational joints to 3-DOF spherical joints.
\end{enumerate}
These improvements aim to reduce motion retargeting errors and enhance the naturalness and fluidity of character movements. The improved model is applicable to most scenarios.

\subsection{Environment Construction}
We utilized object assets from the 3D-Front scene dataset to construct three long-horizon task environments, each designed for distinct experimental scenarios. Environment LH1 is composed of walls, boxes, platforms, and chairs. Environment LH2 incorporates boxes, tables, toy cars, sofas, and television cabinets. Environment LH3 encompasses boxes, tables, toy cars, platforms, nightstands, beds, cushions, and wardrobes. Each environment is configured to support different task specifications.

 We implement standardized transformation and multi-dimensional data extraction for 3D mesh models, providing comprehensive object descriptions for robotic interaction tasks.

\textbf{In the mesh standardization processing stage}: We first load the original normalized models, then apply a specific 90-degree rotation transformation matrix to adjust the coordinate system orientation, followed by calculating the bounding box and repositioning the model at the origin, ensuring all processed objects have a unified spatial reference framework and standardized geometric representation.
\begin{figure}[t]
    \centering
    \includegraphics[width=0.4\textwidth]{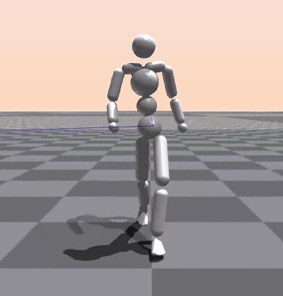}
    \caption{The humanoid we use.}
    \label{fig:111}
\end{figure}
\textbf{In the multi-dimensional feature data generation stage}: We perform bounding box dimension calculations and set standard orientation vectors based on the standardization, use ray tracing techniques to generate $128 \times 128$ resolution height maps to capture fine geometric information of object surfaces, and simultaneously calculate two key target interaction positions based on the height map data—we set the sitting position 0.1 units above the surface and the climbing position directly at the surface height.
\begin{figure*}[t]
    \centering
    \includegraphics[width=1.0\textwidth]{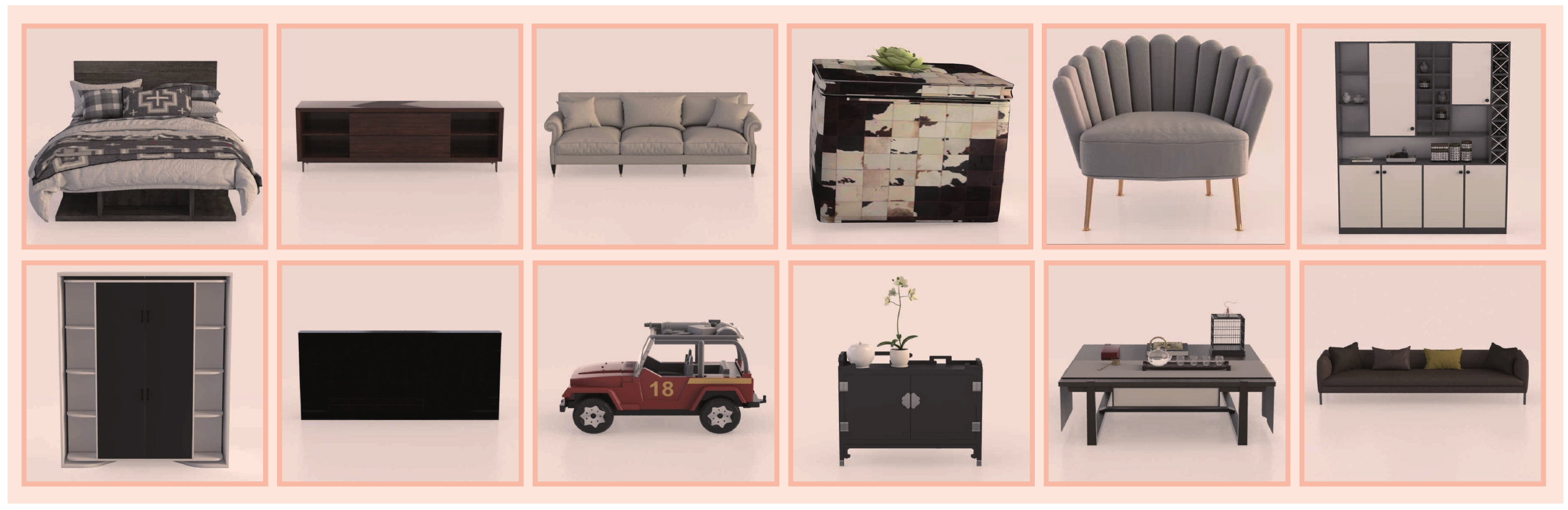}
    \caption{Object assets extracted from 3D-FRONT (3D Furnished Rooms with layOuts and semaNTics)}
    \label{fig:your_label}
\end{figure*}
\textbf{In the complete data output stage}: We generate four types of output data: processed standardized meshes, JSON configurations containing bounding box dimensions and target positions, numerical data and visualization images of height maps, and comprehensive visualization models integrating original models, position marker spheres, bounding box wireframes, and ground planes.

Through the entire processing pipeline, we convert original 3D models into standardized, structured data formats suitable for robotic sitting planning, object climbing, or other spatial interaction tasks, providing a complete geometric and semantic information foundation for our subsequent algorithm development and simulation.
\begin{table*}[htbp]
\centering
\caption{Task Configuration Overview}
\label{tab:task_config}
\resizebox{\textwidth}{!}{
\scriptsize
\begin{tabular}{|l|c|c|c|c|c|c|c|c|c|c|c|c|}
\hline
\multirow{2}{*}{\textbf{Task}} & \multicolumn{4}{c|}{\textbf{Scene Objects}} & \multicolumn{5}{c|}{\textbf{Skills}} & \multicolumn{3}{c|}{\textbf{Complexity}} \\
\cline{2-13}
& \textbf{Total} & \textbf{Boxes} & \textbf{Furniture} & \textbf{Dynamic} & \textbf{Traj} & \textbf{Carry} & \textbf{Climb} & \textbf{Sit} & \textbf{Multi-Traj} & \textbf{Skills} & \textbf{Traj Pts} & \textbf{Targets} \\
\hline
\textbf{LH1} & 5 & 1 & 1 & 1 & {\color{green!70!black}\checkmark} & {\color{green!70!black}\checkmark} & {\color{green!70!black}\checkmark} & {\color{green!70!black}\checkmark} & {\color{red}\texttimes} & 4 & 4 & 2 \\
\hline
\textbf{LH2} & 10 & 2 & 7 & 3 & {\color{green!70!black}\checkmark} & {\color{green!70!black}\checkmark} & {\color{red}\texttimes} & {\color{green!70!black}\checkmark} & {\color{green!70!black}\checkmark} & 4 & 12 & 2 \\
\hline
\textbf{LH3} & 11 & 2 & 8 & 3 & {\color{green!70!black}\checkmark} & {\color{green!70!black}\checkmark} & {\color{green!70!black}\checkmark} & {\color{green!70!black}\checkmark} & {\color{green!70!black}\checkmark} & 5 & 16 & 3 \\
\end{tabular}
}

\centering
\resizebox{\textwidth}{!}{
\begin{tabular}{|l|c|c|c|c|c|c|c|c|c|c|c|}
\hline
\multirow{2}{*}{\textbf{Task}} & \multicolumn{4}{c|}{\textbf{Environment Features}} & \multicolumn{4}{c|}{\textbf{Object Positions}} & \multicolumn{3}{c|}{\textbf{Sampling Config}} \\
\cline{2-12}
& \textbf{Rotation} & \textbf{Rooms} & \textbf{Vehicle} & \textbf{Electronics} & \textbf{Start X} & \textbf{Start Y} & \textbf{Target X} & \textbf{Target Y} & \textbf{Sources} & \textbf{Obj Idx} & \textbf{Trajs} \\
\hline
\textbf{LH1} & {\color{green!70!black}\checkmark} & {\color{red}\texttimes} & {\color{red}\texttimes} & {\color{red}\texttimes} & 10.0 & 4.0 & 0 & -1.8 & 4 & 2 & 1 \\
\hline
\textbf{LH2} & {\color{green!70!black}\checkmark} & {\color{green!70!black}\checkmark} & {\color{green!70!black}\checkmark} & {\color{green!70!black}\checkmark} & 11.5 & 0.0 & 15 & 11 & 4 & 2 & 2 \\
\hline
\textbf{LH3} & {\color{green!70!black}\checkmark} & {\color{green!70!black}\checkmark} & {\color{green!70!black}\checkmark} & {\color{green!70!black}\checkmark} & 11.5 & 0.0 & 15 & 11 & 5 & 3 & 2 \\
\hline
\end{tabular}
}
\end{table*}

\subsection{Task Configuration}

Based on our established long-horizon task scenarios, we designed three tasks with increasing complexity to progressively validate the effectiveness of multi-skill composition.

\subsubsection{LH1: Basic Four-Skill Composition}. LH1 represents a relatively simple long-horizon task designed to validate fundamental four-skill combinations. The scene consists of five core objects: a carriable box positioned at [4.5, -4.0, 0.35], a static chair at [0, 0, 1.46] with -1.57 radians rotation, a base platform, and two boundary walls located at [6, -3.25, 1.5] and [6, 3.25, 1.5] respectively. The execution plan follows [``traj'', ``carry'', ``climb'', ``sit''], requiring the agent to navigate along a predefined trajectory from the starting point [10.0, 4.0] to an intermediate position, pick up and carry the box to the target position [0, -1.8, 0.35], climb to a specified height, and finally sit on the chair to complete the task. We configure target object indices as [0, 0, 0, 1] with sampling sources [``traj\_0'', ``tarpos\_0'', ``scene\_0'', ``scene\_1''].

\subsubsection{LH2: Complex Indoor Scene Navigation} LH2 constitutes a sophisticated indoor scenario that simulates realistic residential interactions. We construct a rich indoor environment with ten objects: two carriable boxes positioned at [11, 11, 0.5] and [15, 18.2, 0.5], furniture including a bed ([15, 20, 1.5]), a table ([15, 13, 0.3]), nightstands ([13.7, 21, 1.7] and [16.5, 21, 1.65]), sofa ([18, 17, 0.35]), TV stand ([18, 13, 0.35]), television ([18, 13, 1.5]), along with a car ([13.2, 16, 0.5]) and a support platform. The execution plan [``traj'', ``carry'', ``traj'', ``sit''] incorporates dual-trajectory navigation: traversing trajectory 0 from [11.5, 0.0] to [11.5, 8.0], carrying the first box to target position [15, 11, 0.35], navigating trajectory 1 from [14, 11.0] to [18.0, 14.4], and sitting on the sofa to complete the task. Target object indices are [0, 0, 0, 8] with sampling sources [``traj\_0'', ``scene\_0'', ``traj\_1'', ``scene\_8''].

\subsubsection{LH3: Comprehensive Five-Skill Integration}. LH3 represents the most complex indoor scenario designed to validate comprehensive five-skill composition capabilities. We extend the environment to eleven objects by adding a second box at [15, 18.2, 0.43] to LH2's configuration, creating a more intricate interaction scenario. The execution plan [``traj'', ``carry'', ``traj'', ``climb'', ``sit''] implements the complete skill sequence: initial trajectory navigation, object manipulation, complex path navigation (trajectory 1 contains twelve waypoints from [14, 11.0] to [15.0, 16.0]), climbing skill execution, and final sitting completion. Target object indices are configured as [0, 0, 0, 1, 2] with sampling sources [``traj\_0'', ``tarpos\_0'', ``traj\_1'', ``scene\_1'', ``scene\_2''], supporting more sophisticated multi-object interactions.

These three tasks follow a progressive complexity pattern, systematically validating multi-skill composition effectiveness in long-horizon tasks from LH1's basic four-skill verification through LH2's complex scene navigation to LH3's comprehensive five-skill integration.

\subsubsection{Task Metrics}
Based on the task configuration table, the three long-horizon tasks (LH1-LH3) demonstrate a progressive complexity design for validating multi-skill composition effectiveness.

\textbf{Scene Complexity:} The environments progress from LH1's basic 5-object setup to LH3's comprehensive 11-object indoor scenario, incorporating carriable boxes, various furniture pieces (chairs, beds, tables, sofas, TV stands), and dynamic elements like vehicles.

\textbf{Skill Integration:} The progression is systematic—LH1 validates fundamental four-skill composition (trajectory, carry, climb, sit), LH2 introduces complex indoor navigation with dual-trajectory execution, and LH3 achieves complete five-skill integration, including multi-trajectory navigation capabilities.

\textbf{Task Complexity Metrics:} This reflects the progression: LH1 requires 4 skills across 4 trajectory points targeting 2 objects, while LH3 requires 5 skills across 16 trajectory points targeting 3 objects.

\textbf{Environmental Features:} The sophistication advances from LH1's simple boundary-wall setup to LH2 and LH3's realistic residential environments featuring room structures, vehicles, and electronics.

\textbf{Execution Plans:} These follow increasingly complex sequences - LH1's basic ``traj-carry-climb-sit'' pattern, LH2's dual-trajectory indoor navigation simulation, and LH3's comprehensive skill sequence validation, creating a systematic framework that progresses from basic skill verification through complex scene navigation to complete multi-skill integration testing.

\begin{table}[htbp]
\centering
\caption{Environment \& Simulation Parameters}
\begin{tabular}{|p{4cm}|p{3cm}|}
\hline
\textbf{Parameter Name} & \textbf{Value} \\
\hline
Parallel Environments & 4096 \\
\hline
Episode Length & 1200 steps (40s) \\
\hline
Control Frequency & 30Hz \\
\hline
Sub-steps & 2 steps \\
\hline
Environment Spacing & 5m \\
\hline
Physics Engine & PhysX \\
\hline
Solver Type & TGS (Type 1) \\
\hline
Position Iterations & 4 iter \\
\hline
Contact Offset & 0.02m \\
\hline
Static Friction & 1.0 \\
\hline
Dynamic Friction & 1.0 \\
\hline
\end{tabular}
\label{tab:env_sim_params}
\end{table}

\begin{table}[htbp]
\centering
\caption{Neural Network Parameters}
\begin{tabular}{|p{4cm}|p{3cm}|}
\hline
\textbf{Parameter Name} & \textbf{Value} \\
\hline
Transformer Layers & 4 layers \\
\hline
Attention Heads & 8 heads \\
\hline
Base Feature Dimension & 64D \\
\hline
Task Observation Space & [128, 96, 112, 144]D \\
\hline
Adapter Units & [1024, 512]D \\
\hline
\end{tabular}
\label{tab:nn_arch_params}
\end{table}

\begin{table}[t]
\centering
\caption{Training \& Reward Parameters}
\begin{tabular}{|p{4cm}|p{3cm}|}
\hline
\textbf{Parameter Name} & \textbf{Value} \\
\hline
\multicolumn{2}{|c|}{\textbf{Training Strategy}} \\
\hline
AMP Observation Steps & 10 steps \\
\hline
Mixed Initialization Prob & 0.5 \\
\hline
State Initialization & Default mode \\
\hline
Max Transition Steps & Train: 60, Test: 20 \\
\hline
Success Threshold & 0.3m \\
\hline
IET Enable & Last subtask only \\
\hline
Task-specific Discrimination & Enabled \\
\hline
\multicolumn{2}{|c|}{\textbf{Reward Function}} \\
\hline
Power Reward & Enabled, coeff: 0.0005 \\
\hline
Trajectory Failure Distance & 4.0m \\
\hline
Termination Height & 0.15m \\
\hline
Dynamic Object Speed Penalty & Coeff: 1.0, Threshold: 1.5 \\
\hline
Decoupling Mask Strength & 0.3 \\
\hline
\end{tabular}
\label{tab:train_reward_params}
\end{table}

\begin{table}[t]
\centering
\caption{Data Generation Parameters}
\begin{tabular}{|p{4cm}|p{3cm}|}
\hline
\textbf{Parameter Name} & \textbf{Value} \\
\hline
\multicolumn{2}{|c|}{\textbf{Height Map Perception}} \\
\hline
Use Height Map & Enabled \\
\hline
Cube Side Length & 2.0m \\
\hline
Grid Points & 25$\times$25 \\
\hline
Grid Spacing & 0.1m \\
\hline
Field of View Spacing & 1.0m \\
\hline
Camera Height & 10.0m \\
\hline
\multicolumn{2}{|c|}{\textbf{Trajectory Generation}} \\
\hline
Trajectory Sample Points & 10 points \\
\hline
Sampling Time Step & 0.5s (10Hz) \\
\hline
Velocity Range & 1.4-1.5 m/s \\
\hline
Maximum Acceleration & 2.0 m/s$^2$ \\
\hline
Sharp Turn Probability & 0.02 \\
\hline
Sharp Turn Angle & 1.57 rad (90$^{\circ}$) \\
\hline
\end{tabular}
\label{tab:data_gen_params}
\end{table}

\subsection{Parameter configuration}
\subsubsection{Environment and Simulation Parameters }
DETACH controls the fundamental simulation setup with 4096 parallel environments, each episode of 1200 steps (40 seconds). Control commands are executed at a 30Hz frequency with 2 physics simulation substeps between controls. We spatially separate environment instances by 5 meters. The PhysX \cite{nvidia_physx_online} physics engine uses TGS (Type 1) solver with 4 position constraint iterations, 0.02m contact detection offset, and both static and dynamic friction coefficients set to 1.0.
\subsubsection{Neural Network Parameters}
The neural network architecture employs a 4-layer Transformer with an 8-head multi-head attention mechanism. Base feature vectors are 64-dimensional, while different tasks have varying observation space dimensions: [128, 96, 112, 144]. Network adapters use hidden layer dimensions [1024, 512] to handle task-specific adaptations.
\subsubsection{Training and Reward Parameters}
The The training strategy utilizes Adversarial Motion Priors (AMP) with 10-step historical observations and skill selection based on specified probability distribution [0.1, 0.1, 0.2, 0.1, 0.1, 0.05, 0.0, 0.05, 0.1, 0.1, 0.1]. We adopt the mixed initialization strategy with 0.5 probability using the default state initialization mode. Task transitions allow a maximum of 60 steps during training and 20 steps during testing, with success determined by a 0.3 m distance threshold. Interactive Early Termination (IET) is enabled only for the final subtask with task-specific discrimination activated. The reward function includes power consumption weighting (coefficient 0.0005), trajectory tracking failure at 4.0m distance, episode termination at 0.15m height, dynamic object speed penalty (coefficient 1.0, threshold 1.5), and decoupling mask strength of 0.3
\subsubsection{Data Generation Parameters}
Height map perception is enabled with a 2.0m cube coverage area using a 25×25 grid resolution (0.1m spacing between adjacent points). The field of view is set to 1.0m with the camera positioned at 10.0m height for data acquisition. Reference trajectories are generated with 10 sampling points at 0.5s intervals (2Hz frequency). Trajectory velocity ranges from 1.4 to 1.5 m/s with a maximum acceleration of 2.0 m/s². Sharp turns occur with 0.02 probability, creating 1.57 radian (90°) angle changes when triggered.

\subsection{Failure Analysis}

We conducted a comprehensive analysis of failure modes across different methods, summarized by task categories below:

\textbf{LH1 Task}
The humanoid agent exhibits failures across multiple execution phases. Initially, spawn point configurations present challenges due to random orientation initialization---specifically, when the humanoid initializes at $-90°$, the $180°$ angular deviation from the target trajectory causes instability during turning maneuvers, frequently resulting in falls. During object manipulation phases, the agent fails to successfully grasp and lift boxes due to insufficient grip strength or improper contact dynamics. In addition, climbing sequences are prone to failure due to foot placement errors that lead to missed footholds and subsequent falls.

\begin{figure*}[h]
    \centering
    \includegraphics[width=1.0\textwidth]{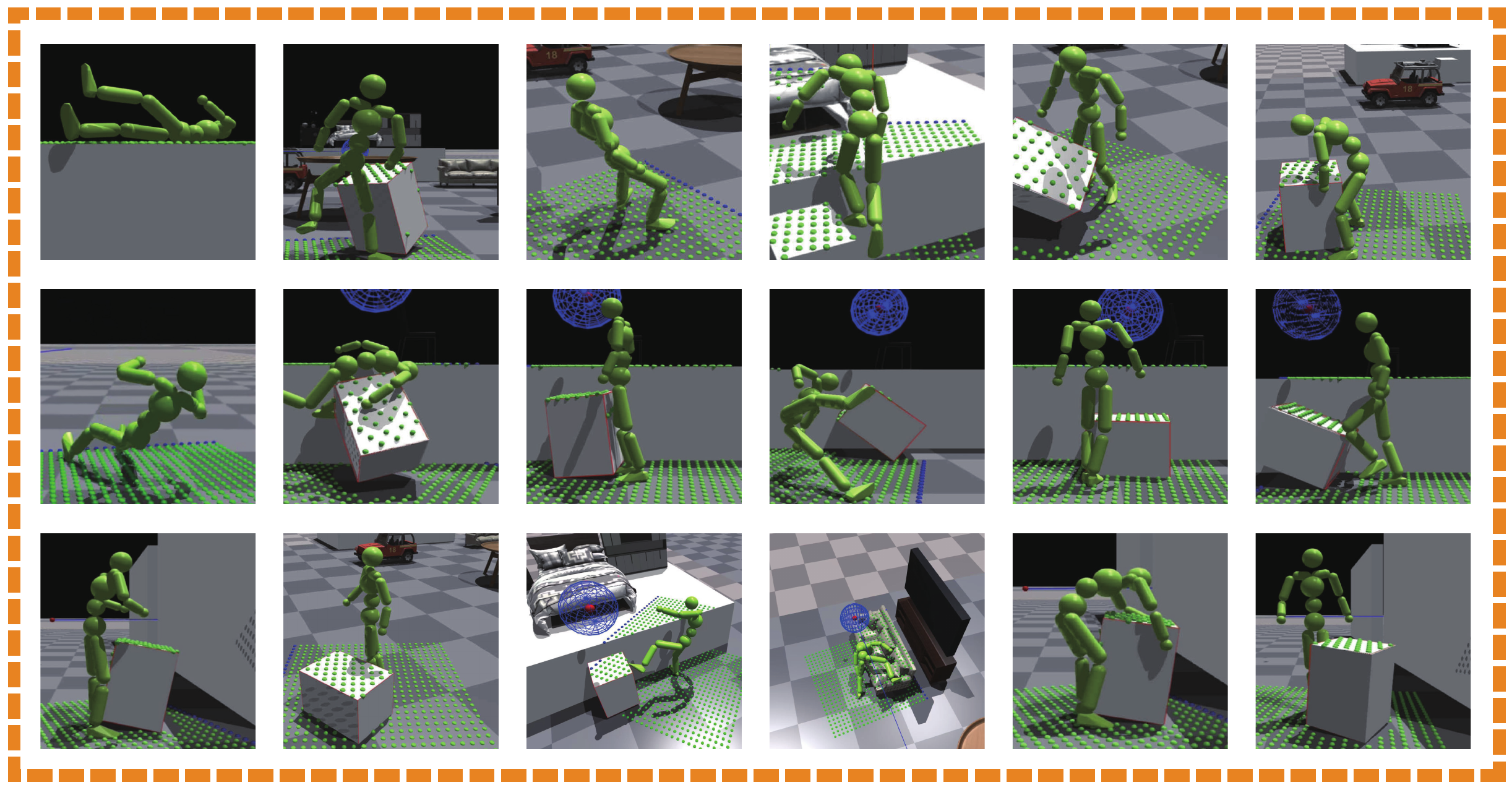}
    \caption{Negative Results Presentation}
    \label{fig:pdf_insert}
\end{figure*}

\textbf{LH2 Task}
Box carrying operations frequently fail due to inadequate lifting capabilities, causing the agent to remain stationary. Subsequently, detection failures occur when the box is not properly positioned within the target zone, preventing trigger activation for subsequent actions and resulting in deadlock states. During locomotion phases, collisions with static environmental objects can cause destabilization and falls. Chair sitting behaviors are compromised by scale mismatches in the furniture models, leading to improper seating postures and misalignment.

\textbf{LH3 Task}
Climbing maneuvers exhibit similar failure patterns to LH1, with foot placement inaccuracies resulting in falls. Bed-sitting tasks present unique challenges due to the geometric constraints and physical properties of the bed model, preventing successful completion of proper sitting behaviors in the conventional sense.

These failure modes highlight the critical need for improved contact dynamics, environmental awareness, and robust motion planning in humanoid robot control systems.

\subsection{Methodological Supplement}
In this section, we supplement the main methodology with additional details that were omitted from the primary exposition.

\subsubsection{Progressive Training Protocol}
To maintain the specialized characteristics of modules during joint training, we designed three key regularization constraints:

\textbf{Feature Decoupling Regularization} Reinforces independence by minimizing mutual information between environmental features and self-features:
\begin{equation}
\mathcal{R}_{decouple} = \|\text{Corr}(z_{env}, z_{self})\|_F^2
\end{equation}
where $\text{Corr}(\cdot, \cdot)$ computes the feature correlation matrix, and $\|\cdot\|_F$ denotes the Frobenius norm.

\textbf{Temporal Consistency Constraint} Applies temporal smoothness constraints to the output of the self-state encoder:
\begin{equation}
\mathcal{R}_{temporal} = \sum_{t=1}^{T-1} \|z^{t+1}_{self} - z^t_{self}\|_2^2
\end{equation}

\textbf{Semantic Preservation Constraint} Ensures that encoded features maintain original semantic information through reconstruction loss:
\begin{multline}
\mathcal{R}_{semantic} = \alpha \|\text{Decoder}_{env}(z_{env}) - obs^t_{env}\|_2^2 \\
+ \beta \|\text{Decoder}_{self}(z_{self}) - obs^t_{self}\|_2^2
\end{multline}

\clearpage

\end{document}